Author:

# Aryia Dattamajumdar

Project Title:

# An early warning AI-powered portable system to reduce workload and inspect environmental damage after natural disasters

*Advisor: Professor William Keat, Union College, Schenectady, New York*

This research did not involve testing with non-human vertebrates

### Acknowledgments:

All experiments were done in the home environment and in my garage lab. I appreciate the financial resources provided by my parents to procure experimental devices. I appreciate the time and feedback provided by officers of the Sunnyvale Fire Department, Sunnyvale, California.

Despite the supreme human achievement of domesticating fire, taming fires remains an act relying on human ingenuity. *How can we make search and rescue operations safer for fire-fighters?*

## Abstract:

1.3 million household fires, 3,400 civilian deaths, and 23 billion dollars in damage, a fire department is called to respond every 24 seconds. Many firefighters are injured during search and rescue operations due to hidden dangers. Additionally, fire-retardant water runoff pollution can threaten human health. My goal is to develop a system to monitor calamity-induced environment damage to provide early-intelligence to incident-commanders.

I have developed a multi-spectral sensing system to inspect air and water quality for safer and accessible hazardous environment operations. Key components include a) drone mounted with four sensors (gas sensors, thermal camera, GPS sensor, visual camera) and wireless communicator for inspection, b) AI-powered computer vision base-station to identify targets, c) low-cost, portable, spectral water quality analyzer and d) robotic retriever.

The prototype demonstrates the potential for safer and more accessible search and rescue operations for fire-fighters and scientists. The gas sensor could identify thick smoke situations (thresholds > 400). The visual and thermal cameras detected hidden hot objects and sent images to AI-powered analyzer to identify and localize target with rescue GPS coordinates for robotic retrieval. Water quality was analyzed with spectral signatures to indicate turbidity levels that correlate with potential pollutants (threshold > 1.3). Prototype results were shown to the Sunnyvale fire department and received encouraging feedback. Future goals include monitoring firefighter health and overexertion with smart clothes.

# 1. Engineering project summary

**Background:** The impact of climate change is highly concerning – with record flooding from Hurricane Florence in South Carolina to prolonged fire seasons in California. Additionally, demographic changes with 30%+ increase inhabitation near wildland-urban interfaces are putting unprecedented collaboration challenges between wildland and municipal fire-fighters. Pollution from water runoff from fire-destruction and floods are threatening human health and property. Many fire-fighters were injured during disaster search & rescue operations due to hidden dangers such as poisonous gases and hidden hot objects. During my research, fire and public safety officials have told me that an early warning system that guides time-sensitive efforts of incident commanders coordinating search and rescue efforts during natural disasters and water quality monitoring would be very valuable. Additionally, Incident Commanders are highly concerned about the physical stress and injuries of the front-line environment protectors.

**Engineering goal:** My goal is to develop an early warning AI-powered portable system that can monitor environmental damage during natural disasters. Both air and water quality needs to be monitored with the aim of providing early intelligence to incident commanders to plan search and rescue operations. Additionally, intelligence on the physical stress of front-line environment protectors will guide incident commanders on optimal resource planning.

**Design, constraints, and prototyping:** A multi-spectral sensing system to monitor air and water quality that can operate in dangerous environments for safer and accessible rescue operations was developed. The prototype has 1) search and inspection multi-sensor drone, 2) AI-powered intelligent base station, 3) a low-cost portable spectral water quality inspection system and 4) rescue system based on robotic crawler and hand. The search module has 4 drone-mounted sensors: gas sensors, thermal camera, GPS sensor, visual camera, and wireless communicator. The gas sensor monitors smoke situations while thermal camera detects hidden hot objects. Spectral water quality monitoring system analyzes water samples and quickly analyzes turbidity levels indicating potential pollutants such as salt and bacterial contaminant levels. The AI-powered intelligent base station (Nvidia Jetson TX2) receives target images and GPS from the search system. The GPS results of the target location are sent to the retrieval robot.

**Test results:** The prototype demonstrates the potential for safer operations for fire-fighters and scientists monitoring water quality to remotely evaluate the search environment for smoke danger, find the target object and communicate GPS location to the base station. Open-source AI algorithms (imagenet/detectnet) were used to successfully identify and localize fires and other predetermined targets. Target object image and GPS location were transmitted to the rescue system. The gas sensors could identify thicker smoke at thresholds above 400. The portable spectral analyzer could determine water turbidity in an easy-to-use, low-cost manner with a threshold of 1.3 Prototype results were reviewed with the Sunnyvale fire department and received positive feedback.

**Conclusion and future work:** This engineering prototype has met the goals to demonstrate a multi-spectral intelligent robotic search and rescue system for safer and better-coordinated fire-fighting and assessing water quality impact. Future enhancements will a) monitor over-exertion related traumas such as *Rhabdomyolysis* with wearable smart clothes, b) improved retriever guidance algorithms and c) build collaboration mechanisms between drones to cover large operation regions for advance warning to human fire-fighters.

## 2. Background and motivation

Environment protection agencies around the world are charged with protecting our crucial environmental quality including air quality and water. Climate change intensified many natural disasters in 2018 from hurricanes, wild-fires, and floods [1]. The prolonged fire-fighting season has stressed the firefighters in recent years.

*How can we make our search and rescue operations safer to help our dedicated environment protectors?*

To study this problem, I reached out to the Sunnyvale Department of Public Safety and the Fire Department. They summarized their challenges in four areas:
1) Safety hazards in fighting fire with limited information on the fire,
2) Coordination of fire-fighting resources especially as wild-land and municipal fire-fighters collaborate in wild-land/urban interfaces,
3) Limiting the downstream impact of fire-fighting on the environment such as water pollution with runoffs, and
4) Early warning of fire-fighter physiological stress due to overexertion in the field

### 2.1. Safety challenges of fire-fighting
1.3 million house-hold fires, 3400 civilian deaths, $23 Billion in damages – a fire department is called to respond every 24 seconds [2]. Despite the supreme human achievement of domesticating fire [3], taming fires continues to rely on human ingenuity. 450 fire-fighters made the ultimate sacrifice in 9/11/2001 NYC terrorist attack. In 2017, 52 of 87 fire-fighter deaths were due to overexertion and 48 who died were volunteers [4]. Rescue operators reportedly fell into septic tanks [5] and canine helpers had severe paw burns [6]. Fire-fighting is often delayed for a lack of local experts. Many victims are injured while searching for their beloved pets as fires progress [7]. These victims were often surprised by poisonous gases, HAZMAT situations, and hidden hot objects, while confined spaces challenged firefighter's access.

### 2.2 Coordination of resources with urban encroachment of wild-lands
Urban encroachment and climate change have packed a mighty punch for Californians in 2018 with wildfires ranging from Southern California's Thomas Fire to the destruction of the town of Paradise in Northern California with at-least 88 deaths. US Forest Service records show a 34% rise in the number of homes in wildland-urban interfaces in California from 1990 to 2010. Similar data exists for Colorado, Montana, and Idaho. California has also been hit by droughts leading to longer fire seasons and significant over-exertion for the firefighters. This combination of climate change, drought, and urban encroachment are leading to a significant rise in the loss of human life and property. An additional challenge is that responsibility for fire-fighting can be unclear between wildland and municipal fire-fighters in these urban-wildland interfaces. The need for collaboration is ever greater than before [1].

### 2.3 Impact of fire-fighting on water pollution
Dumping large quantities of fire retardants on wild-fires bordering urban habitats raises water pollution and sustainability challenges. Fire-suppression with fire-retardants can have a deleterious effect on water resources and aquatic flora/fauna by increasing erosion, sedimentation, turbidity, and chemical contamination – an example being the infamous shutdown of Denver's water supply with the Buffalo creek fire in Colorado in 1996 [8-10]. Organo-halogens such as brominated and chlorinated compounds are often used in many flame retardants. Many of these compounds such as PDBEs are known for environmental toxicity and persistence [11]. While PDBEs have been phased out from newer flame retardants, they are still present in many older homes that have been impacted by wild-fires. The dioxins and furans released during the burning of halogenated fire retardants can runoff into water supplies with storm runoffs [12-14]. Studies have shown that chemical contamination of fire retardant foam can negatively impact aquatic fauna [15]. The water contaminants are typically analyzed by sophisticated and expensive equipment such as gas chromatography or mass spectroscopy [16].

Incident commanders have told me that having low-cost portable equipment to measure water sedimentation and turbidity levels would be helpful in downstream actions such as runoff management. Turbidity poses indirect threats to human health by encouraging microbial production and increasing the risks of contracting infections for people who come in contact with untreated water. Excessive amounts of sediment in a water supply may stress filtration systems at water treatment facilities.

**2.4 Physiological stress of over-exertion**
Fire-fighting is one of the most strenuous and mentally exhausting occupations. Fire-fighters have to analyze a large variety of sensory inputs such as assessing the environment, movement of people and anticipate hazards. The weight of the fire-fighting protective clothing and breathing gear to be 45-75 pounds [17]. This overload can impact their metabolic and thermo-regulatory balance, fatigue and lead to injuries. They have suffered nearly eleven thousand injuries a year due to slips, trips and falls accounting for 25% of all injuries [18]. While fire-fighters typically start their career in great physical condition, age-related changes in body-fat and decreasing lean mass can lead to a higher risk of injuries [19]   Most assessments of firefighter physiology is done under controlled laboratory conditions or in simulated environments due to the nature of the measurement equipment [20-22]. Prolonged muscle and physiological stress during wild-land firefighting can lead to conditions such as *Rhabdomyolysis* wherein the muscles breakdown and impact kidney function [23]. 19 confirmed cases of *Rhabdomyolysis* have been reported in 2008-2015 [24].

# 3. Engineering goals

My goal is to develop an early warning AI-powered portable system that can monitor environmental damage during natural disasters. Both air and water quality needs to be monitored with the aim of providing early intelligence to incident commanders to plan search and rescue operations. Additionally, intelligence on the physical stress of front-line environment protectors will guide incident commanders on optimal resource planning.

There are two key engineering goals:
   1) Develop an intelligent search and rescue system based on multi-spectral sensing to enable safer, more accessible operations.
   2) Develop a portable system to measure turbidity in water samples to enable incident commanders to appropriately plan storm runoffs of fire retardants

# 4. Design criteria

The prototype will have four modules as shown in Figure 1:

1) Search and inspection drone with multispectral sensors
2) AI-powered intelligent base station
3) Portable spectral system for quick measure of water turbidity
4) Rescue retrieval system

Drone mounted sensors will evaluate the search region for dangerous gases and hidden hot objects with gas sensors, visual and thermal images. GPS location and images of rescue targets will be sent wirelessly to the AI-powered analyzer station to localize the target. The identified target's images and GPS locations will be sent to the retrieval robot for rescue. The schematic is shown in Figure 1.

To measure water turbidity in the field, I was inspired by my childhood observation that ocean water is blue but river water is relatively clear as in waterfalls. Blue light is not absorbed by water but is absorbed by common salt. The spectra of water and salt are shown in figure 2. To measure turbidity that can represent microbial growth, I will measure the optical density of the water sample.

A future goal is to develop a real-time light wearable sensor to monitor muscle fatigue and physiological stress that can be incorporated into firefighters' personal protective gear. Physiological performance can be monitored by measuring muscle activity with surface EMG as shown in figure 3(a) [27] while heart rate can be monitored by EKG [28] as shown in figure 3(b). Brain activity could be measured by recording EEG signals [29] as shown in figure 3(c). Real-time, low-cost, miniature ECG, EMG and EEG sensors will be used to measure the physiological stress of the firefighters.

## 5. Design constraints

Fire departments have used drones for 6 major purposes i.e. scene monitoring, search and rescue, pre-fire training, post-fire disaster assessment, wildland fire response and emergency deliveries [30, 31]. Fire-fighters have long used manned aircraft with infrared sensors to detect and guide rescue crews [32]. Goldman Sachs has estimated a fire-fighting drone market to be worth $881 million globally [33]. However, professional fire-fighting can be interrupted by hobbyists flying drones to capture aerial videos [34]. Urban fire-fighters (e.g. NYC, Menlo Park) have used expensive drones ($85,000+) [35] such as Matrice 210, Zenmuse XT and Z30 since 2017 to provide aerial footage with visual and night-cameras. These drones are typically human monitored with special operation permits [36, 37]. Recently, Parrot has released drones costing $5000 for fire-fighters [38]. However, low-cost smart fire-fighting tools to help citizen volunteers are still many years away.

Real-time fire-fighting field data is very limited and actions are often based on the incident commander's experience [39]. NIST and DHS NGFR teams [40, 41] have identified key challenges in smart fire-fighting:

### 5.1 Decision making
Fire-fighters need timely and accurate information. Often, a deluge of reports is received that need investigation prior to dispatch of professional teams. Standard maps often don't work well on disaster sites as they need to be updated in real-time for hazards. JPL's AUDREY project has attempted to use cloud-based AI for decision-making [42].

### 5.2 Overwhelmed communication systems
Infrastructure often breaks down at disaster sites. Localized wireless communication independent of on-ground ICT is needed. Cloud-based localization and tracking analytics (e.g. NEON Personnel Tracker [43] and Precision Location and Mapping System [44]) may have limited viability.

**5.3 Environmental cost of fire-retardant**
Manned aircraft fight fire by dropping massive amounts of fire-extinguishing fluids. Burning of these organohalogens can lead to dioxins and furans which can runoff into the water supply during storms. If incident commanders could monitor real-time water quality, they could build better barriers to limit the environmental contamination of fire-retardants.

**5.4 Monitoring firefighter's physical condition**
Cardiac failure and *Rhabdomyolysis* due to overexertion with carrying heavy fire-fighting protective gear over long periods of time are major causes of firefighter deaths. How can we give fire-fighters real-time feedback whether they have exceeded the physiological safe operation zone? While current efforts (e.g. ZEPHYR performance system [45-46]) have monitored heart and respiratory rate, muscle and brain activity monitoring could provide additional insights.

Fire-fighting response requires firefighters to make crucial context-dependent decisions [47]. Discussions with Sunnyvale fire department were that they have found digital maps, indoor navigation and local environmental conditions reporting especially for HAZMAT situations very valuable. Many remain skeptical of drones, on-ground robots and intelligent clothing [48].

## 6. Building the prototype

I have built a prototype that demonstrates a solution to the first three challenges i.e. drone-mounted sensors to evaluate search region with automated decision making with AI-powered algorithms and wireless transmission of data without reliance on-ground infrastructure and a portable spectral analyzer for providing quick, easy-to-use the results of water turbidity. The prototype has 4 major subsystems (figure 1) i.e.

1) Search and inspection of multispectral drone
2) AI-powered intelligent base station
3) Rescue system
4) Portable spectral module to test water quality

A drone searches for a target with a novel sensor-pack. Data are sent wirelessly to an AI-powered base station to automatically identify the target. Target GPS location is transmitted to rescue robotic retriever. Robotic retriever compares target GPS location with the current location and is guided towards the target while avoiding obstacles.

**6.1 Search and inspection module**
Figure 4 shows the schematic of the search module. Visual and thermal imaging sensors are used to identify the target. GPS provides the target location. Gas sensors evaluate rescue operation viability.

Figure 5 shows the schematic of the gas sensor. Sensitive SnO2 active element identifies smoke and gases (LPG, propane, hydrogen, and methane) [49, 50]. The gas sensor was connected to the Arduino_UNO_R3 board along with buzzer and alarm LEDs (figure 6). As gas levels increase, output voltage increases.

Arduino shield (1sheeld) unlocked sensors in iPhone5 including camera, GPS, accelerometer, magnetometer, and microphone [51, 52]. A low-cost thermal camera (FLIR SEEK) connected to the iPhone5 was used to detect heat signatures [53]. A wireless NRF24L01 module connected to Arduino_UNO_R3 was used to transmit gas sensor readings to the base station. Figure 7 shows the constructed engineering prototype.

Two drones were tested to lift the sensor-pack. Cheerwing X5SW drone has a 3.7V 550 mAh battery and could not lift a sensor-pack. Next, I tested a Force 100 ghost drone with 7.4V, 1800 mAh battery [54] which could easily lift sensor-pack. Figure 8 shows the integrated search module.

## 6.2 AI-powered intelligent base station
Figure 9 shows a schematic of the base station. A wireless board (MakerFocus NRF24L01+PA+LNA) and LCD were connected to the Arduino_MEGA2560 board to receive target image and location and transferred for analyses to Nvidia_JetsonTX2 board [55]. The JetsonTX2 integrated video camera was used to view potential targets (fire engines, fires, dogs). I have used *'imagenet'* and *'detectnet'* AI algorithms on JetsonTX2 to identify and locate targets [56]. The resulting identified object and GPS location are sent wirelessly by the base station to the rescue module. A future goal is to fuse drone sensor data with target identification & location results on an augmented reality display for seamless fire-fighter workflow. Figure 10 shows the constructed base station.

## 6.3 Rescue and retrieval robot
Figure 11 shows the rescue retriever schematic. It has a Devastator$^{TM}$ crawler tank [57] and a six degree of freedom retrieval robotic hand (figure 12) [58]. Robotic tank with Arduino guided motors and guidance system are shown in figure 13. Robotic hand has 6 independently controlled servo motors via I2C protocol using PCA9685 board [59] (figure 14). The robotic tank is GPS guided to the target via iPhone4S sensors unlocked with Arduino 1sheeld. Retrieval target is identified via PixyMon camera while obstacles are avoided with multiple ultrasonic distance sensors (figure 15). Fine control of the robotic hand is LIDAR-guided (figure 16). These sensors have been individually tested but have not been fully integrated into the robotic rescue system at this time. Figure 17 shows the sensor map for the search and rescue system.

## 6.4 Spectral module to measure water turbidity
A block diagram schematic of the module for measuring water turbidity based on spectral changes due to light absorption is shown in figure 18. The device contains a programmable RGB common cathode LED. This LED can be programmed to emit various color lights by combining red, green and blue LEDs in the package. The intensity can be varied based on the amount of current that is sent from an Arduino microcontroller as shown in Figure 19.

The device uses a circuit that has a light detecting resistor (LDR) to sense light transmitted through the solution being tested as shown in figure 19(a). The resistance of the LDR changes depending on the amount of light that falls on it as shown in figure 19(c). The signal is amplified using a transistor. A test LED gives visual results of the amount of light falling on LDR and is useful for tuning and troubleshooting the circuit. A potentiometer is used to tune the transistor circuit based on light/dark resistance of LDR. The amount of light absorbed is detected by a light detecting resistor (LDR) which is connected to an amplifier and a test LED. As a particular color of light is absorbed, less light passes through to LDR. This increases the resistance of LDR which is then converted into an electrical signal and amplified. The results are measured across the transistor using Arduino analog read pin on the Arduino microcontroller and read via the serial controller interface on the laptop.

The prototype of the device was built for testing and is shown in figure 20 with an inside look of the assembled prototype with programmable RGB LED, light-detecting resistor (LDR) amplifier sensor, Arduino microcontroller and USB serial interface to the laptop. Three different LDRs of 3 mm, 5 mm and 7 mm were tried to find the optimal device setup. The sample cuvette was stably and consistently positioned using the cuvette positioner.

## 7. Health and safety

A safe environment was maintained while developing the system in our home environment (garage lab). Precautions taken included:

1) Gas sensor and thermal camera testing: Precautions, as learned in Chemistry laboratory at high school, were observed. This included wearing protective clothing and eye goggles. Match-sticks and incense-sticks were used to simulate smoke under adult supervision. Bucket of water was kept available to douse match-stick, incense-stick, and a small fire. Car exhaust was tested in an open-environment under adult supervision.
2) Electric power: Precautions were taken to keep circuits disconnected from power while working on them.
3) Drone testing: The drone was flown and tested indoors with protective cages on the propellers. Limited outdoor testing was done in a park under adult supervision and direct visual control. In the future, a drone license will be obtained from the FAA under CFR part 107 to develop larger experimental drones.
4) Retrieval system testing: Rescue robot was tested indoors at relatively slow speeds.
5) Fire-testing: Videos of wild-fires and urban house fires were retrieved from YouTube [60, 61]. This enabled testing for fires without being a dangerous environment.
6) The spectral module for measuring water turbidity does not pose any significant risk to the user as it does not use any ionizing radiation or harmful radiation. The RGB LED emits light in the safe visual range. The liquids are placed in a cuvette that is isolated from the electronics through the foam-board enclosure. Common precautions that were taken included being careful about not spilling liquids near the electronics and not looking directly at the RGB LED light source

## 8. Results

Results from unit testing and system integration are summarized below:

### 8.1 Testing gas sensor
The gas sensor was tested in 6 environments to measure smoke levels under adult supervision: Normal home-room environment as control; smoke from match-stick; smoke from aromatic incense-stick; smoke from soot-less tea-candle; exhaust from an older car; and exhaust from a newer car. Three measurements were taken for each scenario as shown in table 1 and figure 21. A threshold of 200 can distinguish high smoke situation from a low smoke situation with this gas sensor. In the future, the sensor will be calibrated with known gas concentrations.

### 8.2 Testing thermal camera
Thermal camera was tested in four situations to detect hidden/invisible situations i.e. orange hidden in dark plastic bag; human arm in a dark plastic bag; cars in night environment; and orange obscured by smoke. Figure 22 shows the results of thermal camera images detecting objects missed in visual images. Additionally, hot exhaust could identify a car better than night-time visual images. Figure 23 shows that thermal camera can detect targets like orange obscured by smoke or fire in visual images.

### 8.3 Testing AI-based target object identification and location
JetsonTX2 integrated camera was used to view images of multiple dog breeds, forest fires [60] and fire engines [61] from YouTube. The **imagenet** algorithm identified dog breeds (figure 24), fire-engines (figure 25a) and fires (figure 27c) with 60-95% confidence. While it detected fire-engine well in side-view, it confused fire-engine as ambulance or moving van in front-view (figure 25b).
The **detectnet** algorithm was used to localize target. It was good at localizing targets such as dogs and people (figure 26). However, it also had a number of false-positive identifications.

### 8.4 Testing robotic retriever parameters
The robotic tank was powered by 4 AA batteries but it moved very slowly and the batteries were soon exhausted. Next, a set of 3 x 3.7V batteries with 3800 mAh capacity were tested and a DC-DC converter was used to step down the voltage to 4-6V range with over 0.5 A current sourced.

To find optimal drive voltage and load-carrying capacity of the robotic tank, a number of tests were done. First, the robotic tank crawler was tested in a "no-load" situation. At voltages below 5.5V, the motors would not turn (only resonant noise was heard). As the voltage was increased from 5.5V to 6.8V using the DC-DC converter, tank speed increased. Results from the test are shown in table 2 and figure 27. Next, the robotic tank was tested with a series of loads ranging from 200 grams to 2 kilograms. As the load was increased from 200g to 2000g, the travel time of the robotic tank reduced as shown in table 3 and figure 28.

The robotic hand servos were set up and controlled to pick-up objects. The ultrasonic sensor and LIDAR parameters were set up. Seamless sensor integration for robotic retrieval and guidance will continue to be developed and tested in the future.

### 8.5 Calibrating the spectral module to detect water turbidity
Three different sensors with 3 mm, 5 mm and 7 mm LDR were built. The distance between the light source and LDR was 4 cm for the sensors with 3 mm LDR and 5 mm LDR while the distance was adjusted to a lower 3 cm for 7 mm LDR sensor as it had very weak response.

Each sensor was individually tuned by adjusting the potentiometer to maximize the range of response to different intensities of blue light for reference salt solution as this was the weakest signal. The intensity of remaining color lights was adjusted lower so that the sensor output was not saturated. The 3 mm LDR sensor was adjusted to tuning value of 235 Ohms, while 5 mm LDR needed a tuning value of 623 Ohms and the 7 mm LDR needed a tuning of 973 Ohms. The sensor with 7 mm LDR had very weak response to blue light compared to other lights. Measurements were taken with a number of samples including plain water, two salt solutions with different concentrations and two sugar solutions with differing concentrations. Goal was to select the sensor which demonstrated most variation across the samples while being most stable with least variance across 5 measurements for a sample.

The analysis of responses to these three sensors is shown in figure 29. The response to the 3 mm LDR sensor showed a high variance in measurements for a given sample and was not suitable. This may be because the sensor would be sensitive to variations in the alignment of the light source, sample cuvette, and LDR. The response to 7 mm LDR was very weak for blue light and weak across other samples too. This sensor was not suitable given its low sensitivity.

The response to the 5 mm LDR sensor was the best setup for differentiating between solutions with various salt and sugar concentrations as variance across samples was low. The 5 mm LDR response showed most variation across sample types for blue light (1.77V - 3.28V) and showed the least variation for yellow light (3.14V - 3.31V). To protect against the possibility that the RGB LED may dim over its lifetime, a ratio of response between sensitive blue light and insensitive yellow light between the test sample and reference water sample was developed as the differentiator. This ratio is called the Blue-to-Yellow Response ratio (BYR) and is shown in equation 1.

$$\text{Differentiator (BYR)} = \frac{\text{Blue}_{sample/water}}{\text{Yellow}_{sample/water}} \tag{1}$$

Further tests were based on developing algorithms for finding differentiating thresholds for BYR ratio.

**8.6 Testing water quality test module**
The device was tested with 7 reference solutions to check responsiveness. Reference solutions of three concentrations of salt and sugar were prepared with 1 teaspoon, 2 teaspoon, 3 teaspoon in 80 ml of water. Measurements were taken with each of the six reference sample solutions and one solution of plain water. Measurement data are in table 4 for salt solutions and table 5 for sugar solutions. The results of BYR ratio analysis are shown in figure 30.

**8.7 Measuring water turbidity:**
Collecting true samples of water runoff in fire-ravaged environment was not feasible due to public safety restrictions. Turbidity poses indirect threats to human health by encouraging microbial production and increasing the risks of contracting infections for people who come in contact with untreated water. As a simulated test, I used coconut water which I knew attracts microbial growth when left at room temperature for several days. Drinking home water was used as the control sample. BYR ratio was measured over four days at 1) 16 hours, 2) 25 hours, 3) 48 hours, 4) 64 hours, 5) 71 hours and 6) 97 hours. The results of the measurements are summarized in table 6 and shown in Figure 31. For the first 24 hours, BYR ratio was 1.26 ± 0.01. After 48 hours, the sample started turning turbid and BYR ratio jumped to 1.37 ± 0.01. There was a clear threshold in the measurement results of 1.3 to detect turbidity.

## 9. Discussion of results

Nature's fury and human-induced disasters leave a trail of destruction of lives and property – be it wild-fires, earthquakes, hurricanes or terrorist attacks. Following such disasters, rescue teams often work in dangerous and unknown environments to save lives and recover dear property. I have developed a multi-spectral sensing and inspection drone with automated target identification, decision-making and robotic retrieval. It can operate in dangerous environments to make search and rescue operations safer and more accessible.

The prototype has three-four key functioning modules: Search and inspection drone with multispectral sensors; AI-powered intelligent base station; Portable water quality spectral analyzer; and Rescue retrieval system based on robotic crawler and hand. The search module has four drones mounted sensors i.e. gas sensor to evaluate the environment, thermal camera to detect hidden objects, iPhone5 unlocked sensors like GPS and visual camera and a wireless communicator. The gas sensor monitors rescue operation viability while visual and thermal cameras detect search targets. The AI-powered intelligent base station receives target images and GPS from the search system and automatically searches for the target. Open-source AI algorithm **imagenet** was used to identify targets with 60-95% confidence, while another AI algorithm **detectnet** was used to localize the target. Target image and GPS location were transmitted to the robotic retriever. The prototype robotic crawler could carry loads of up to 2 kg and navigates by comparing the GPS location of the target with the current GPS location. It has 6 degrees of freedom. LIDAR guides fine motor control for picking up objects with 5mm accuracy. Robot retrieval is guided by a PixyMon texture-based camera while obstacle avoidance is done with 3 ultrasonic sensors. Obstacle avoidance and retriever guidance will continue to be developed in the future.

The water quality analyzer is based on Beer-Lambert and could detect changes in the concentration of solutes such as salt and sugar. It provides a quick, easy-to-use, low-cost way to assess water turbidity (at a threshold of 1.3) empowering incident commanders at natural disaster sites to minimize the environmental impact of fire-retardant runoffs. As the concentration of salt and sugar increased, the BYR ratio increased (though not linearly). As the concentration of solutes such as salt or sugar increases, more light is absorbed by the solution. More the light absorbed, the lesser the light that falls on the LDR. Lesser the light on LDR, the higher the resistance. This change in signal is amplified by the transistor and is visually seen on the test LED. The lesser light on LDR, the lower the test LED glows and higher the voltage across the transistor collector and emitter.

Prototype results were shown to the Sunnyvale fire department. Officer Wilkes was very enthusiastic about the utility of such a system for HAZMAT situations and combating wildfire situations. The prototype could enable safer search and rescue operations for fire-fighters by continuously evaluating the search environment remotely for early warning. Key improvement feedback included searching for small fires early during RED ALERT situations, improving the range of poisonous gases detected during HAZMAT situations and continuously monitoring fire-fighter physiology for over-exertion and *Rhabdomyolysis*. I am continuing to develop a real-time, low-cost wearable sensor to monitor firefighter physiology that can be incorporated into protective gear.

## 10. Limitations

Key challenges faced in building prototype were in integration, software development, and testing environment. To improve reliability, I plan to use a PCB for the drone-based sensors. I had to manually test the image analysis and detection capability of AI-powered Jetson TX2 as the Arduino IDE did not work with upgraded Jetson TX2. In the future, I plan on getting the FAA CFR part 107 license for operating drones for further testing.

Getting stable & consistent results from the water spectral analyzer was initially a challenge as the sample cuvette tended to slide around and would not be at the correct position in front of the LDR. To overcome this issue, I built a cuvette positioner with black foam-board to keep the cuvettes in a stable position. Additionally, the light source of the spectral analyzer was not very stable which was resolved by building a fixture to hold the LED in a fixed position. The stability of the device could be further improved by constructing it with firm materials and replacing the breadboard and wires with a printed circuit board. LDR response was calibrated using a potentiometer. To get the most sensitive response to various color light, the tuning values were different. Since I am using only one potentiometer, the readings were not optimal for each color.

## 11. Future work

### 11.1 Physiology monitoring module

A module to monitor the physiology of fire-fighters to provide real-time exertion levels to incident commanders is being developed currently based on Arduino sensors for EMG [27] and EKG [28] as shown in figure 32. Figure 33 shows a Neurosky Mindwave Mobile 2 module is being investigated for EEG. The EMG, EKG and EEG signals will be transmitted wirelessly to base analyzer station for interpretation by the incident commander as shown in Figure 33.

**11.2 Retriever guidance algorithms**

The retrieval module has been built with multiple sensors that can aid in guidance, obstacle avoidance and fine motor control for the gripper. Preliminary guidance algorithms have been developed as shown in figure 34 and are being tested.

## 12. Conclusions

In conclusion, this engineering prototype has met the project goal to demonstrate a multi-spectral intelligent robotic search and rescue system for safer and better-coordinated fire-fighting and assessing water quality impact. Future enhancement will include 1) calibrate gas sensors with known gas concentrations to identify gas type, 2) seamless integration of 5 modules, 3) build collaboration mechanisms between drones to cover large areas of search and rescue target regions for advance warning to human fire-fighters, and 4) monitoring fire-fighter physiology with wearable EKG, EEG and EEG sensors.